# Committee-Based Sample Selection For Probabilistic Classifiers

**Shlomo Argamon-Engelson** ARGAMON@MAIL.JCT.AC.IL
*Department of Computer Science*
*Jerusalem College of Technology, Machon Lev*
*P.O.B. 16031*
*Jerusalem 91160, Israel*

**Ido Dagan** DAGAN@CS.BIU.AC.IL
*Department of Mathematics and Computer Science*
*Bar-Ilan University*
*52900 Ramat Gan, Israel*

## Abstract

In many real-world learning tasks it is expensive to acquire a sufficient number of labeled examples for training. This paper investigates methods for reducing annotation cost by *sample selection*. In this approach, during training the learning program examines many unlabeled examples and selects for labeling only those that are most informative at each stage. This avoids redundantly labeling examples that contribute little new information.

Our work follows on previous research on Query By Committee, and extends the committee-based paradigm to the context of probabilistic classification. We describe a family of empirical methods for committee-based sample selection in probabilistic classification models, which evaluate the informativeness of an example by measuring the degree of disagreement between several model variants. These variants (the committee) are drawn randomly from a probability distribution conditioned by the training set labeled so far.

The method was applied to the real-world natural language processing task of stochastic part-of-speech tagging. We find that all variants of the method achieve a significant reduction in annotation cost, although their computational efficiency differs. In particular, the simplest variant, a two member committee with no parameters to tune, gives excellent results. We also show that sample selection yields a significant reduction in the size of the model used by the tagger.

## 1. Introduction

Algorithms for supervised concept learning build classifiers for a concept based on a given set of labeled examples. For many real-world concept learning tasks, however, acquiring such labeled training examples is expensive. Hence, our objective is to develop automated methods that reduce training cost within the framework of *active learning*, in which the learner has some control over the choice of examples which will be labeled and used for training.

There are two main types of active learning. The first uses *membership queries*, in which the learner constructs examples and asks a teacher to label them (Angluin, 1988; MacKay, 1992b; Plutowski & White, 1993). While this approach provides proven computational advantages (Angluin, 1987), it is not always applicable since it is not always possible to construct meaningful and informative unlabeled examples for training. This difficulty may be overcome when a large set of unlabeled training data is available. In this case the second type of active learning, *sample selection*, can often be applied: The learner examines many unlabeled examples, and selects only the most informative ones for learning (Seung, Opper, & Sompolinsky, 1992; Freund, Seung, Shamir, & Tishby, 1997; Cohn, Atlas, & Ladner, 1994; Lewis & Catlett, 1994; Lewis & Gale, 1994).

In this paper, we address the problem of sample selection for training a probabilistic classifier. Classification in this framework is performed by a probability-based model which, given an input example, assigns a score to each possible classification and selects that with the highest score.

Our research follows theoretical work on sample selection in the Query By Committee (QBC) paradigm (Seung et al., 1992; Freund et al., 1997). We propose a novel empirical scheme for applying the QBC paradigm to probabilistic classification models (allowing label noise), which were not addressed in the original QBC framework (see Section 2.2). In this *committee-based selection* scheme, the learner receives a stream of unlabeled examples as input and decides for each of them whether to ask for its label or not. To that end, the learner constructs a 'committee' of (two or more) classifiers based on the statistics of the current training set. Each committee member then classifies the candidate example, and the learner measures the degree of disagreement among the committee members. The example is selected for labeling depending on this degree of disagreement, according to some selection protocol.

In previous work (Dagan & Engelson, 1995; Engelson & Dagan, 1996b) we presented a particular selection protocol for probabilistic concepts. This paper extends our previous work mainly by generalizing the selection scheme and by comparing a variety of different selection protocols (a preliminary version appeared as Engelson & Dagan, 1996a).

### 1.1 Application To Natural Language Processing

Much of the early work in sample selection has either been theoretical in nature, or has been tested on toy problems. We, however, are motivated by complex, real-world problems in the area of statistical natural language and text processing. Our work here addresses the task of *part-of-speech tagging*, a core task for statistical natural language processing (NLP). Other work on sample selection for natural language tasks has mainly focused on text categorization problems, such as the works of Lewis and Catlett (1994), Liere and Tadepalli (1997), and McCallum and Nigam (1998).

In statistical NLP, probabilistic classifiers are often used to select a preferred analysis of the linguistic structure of a text, such as its syntactic structure (Black, Jelinek, Lafferty, Magerman, Mercer, & Roukos, 1993), word categories (Church, 1988), or word senses (Gale,

Church, & Yarowsky, 1993). The parameters of such a classification model are estimated from a training corpus (a collection of text).

In the common case of supervised training, the learner uses a corpus in which each sentence is manually annotated with the correct analysis. Manual annotation is typically very expensive. As a consequence, few large annotated corpora exist, mainly for the English language, covering only a few genres of text. This situation makes it difficult to apply supervised learning methods to languages other than English, or to adapt systems to different genres of text. Furthermore, it is infeasible in many cases to develop new supervised methods that require annotations different from those which are currently available.

In some cases, manual annotation can be avoided altogether, using self-organized methods, such as was shown for part-of-speech tagging of English by Kupiec (1992). Even in Kupiec's tagger, though, manual (and somewhat unprincipled) biasing of the initial model was necessary to achieve satisfactory convergence. Elworthy (1994) and Merialdo (1991) have investigated the effect of self-converging re-estimation for part-of-speech tagging and found that some initial manual training is needed. More generally, the more supervised training is provided, the better the results. In fact, fully unsupervised methods are not applicable for many NLP tasks, and perhaps not even for part-of-speech tagging in some languages. Sample selection is an appropriate way to reduce the cost of annotating corpora, as it is easy to obtain large volumes of raw text from which smaller subsets will be selected for annotation.

We have applied committee-based selection to learning Hidden Markov Models (HMMs) for part-of-speech tagging of English sentences. Part-of-speech tagging is the task of labeling each word in the sentence with its appropriate part of speech (for example, labeling an occurrence of the word 'hand' as a noun or a verb). This task is non-trivial since determining a word's part of speech depends on its linguistic context. HMMs have been used extensively for this task (e.g., Church, 1988; Merialdo, 1991), in most cases trained from corpora which have been manually annotated with the correct part of speech for each word. Our experiments on part-of-speech tagging, described in Section 6.5, show that using committee-based selection results in substantially faster learning rates, enabling the learner to achieve a given level of accuracy using far fewer training examples than by sequential training using all of the text.

## 2. Background

The objective of sample selection is to select those examples which will be most informative in the future. How might we determine the informativeness of an example? One approach is to derive an explicit measure of the expected amount of information gained by using the example (Cohn, Ghahramani, & Jordan, 1995; MacKay, 1992b, 1992a). For example, MacKay (1992b) assesses the informativeness of an example, in a neural network learning task, by the expected decrease in the overall variance of the model's prediction, after training on the example. Explicit measures can be appealing, since they attempt to give a precise characterization of the information content of an example. Also, for membership querying, an explicit formulation of information content sometimes enables finding the most informative examples analytically, saving the cost of searching the example space. The use of explicit methods may be limited, however, since explicit measures are generally (a) model-specific, (b) complex, often requiring various approximations to be practical, and (c) depend on the accuracy of the current hypothesis at any given step.

The alternative to measuring the informativeness of an example explicitly is to measure it implicitly, by quantifying the amount of uncertainty in the classification of the example given the current training data. The informativeness of an example is evaluated with respect to models derived from the training data at each stage of learning. One approach is to use a single model based on the training data seen so far. This approach is taken by Lewis and Gale (1994), for training a binary classifier. They select for training those examples whose classification probability is closest to 0.5, i.e, those examples for which the current best model is most uncertain.

In order to better evaluate classification uncertainty with respect to the entire space of possible models, one may instead measure the classification disagreement among a sample set of possible models (a *committee*). Using the entire model space enables measuring the degree to which the training entails a single (best) classification for the example. On the other hand, referring to a single model measures only the degree to which that model is certain of its classification. For example, a classifier with sufficient training for predicting flips of a coin with heads probability 0.55 will always predict heads, and hence will make mistakes 45% of the time. However, although this classifier is quite uncertain of the correctness of its classification, additional training will not improve its accuracy.

There are two main approaches for generating a committee in order to evaluate example uncertainty: the *version space* approach and the *random sampling* approach. The version space approach, pursued by Cohn et al. (1994) seeks to choose committee members on the border of the space of all the models allowed by the training data (the *version space*, Mitchell, 1982). Thus models are chosen for the committee which are as far from each other as possible while being consistent with the training data. This ensures that the models will disagree on an example whenever training on the example would restrict the version space.

The version space approach can be difficult to apply since finding models on the edge of the version space is non-trivial in general. Furthermore, the approach is not directly applicable in the case of probabilistic classification models, where almost all models are possible, though not equally probable, given the training. The alternative is random sampling, as exemplified by the Query By Committee algorithm (Seung et al., 1992; Freund et al., 1997), which inspired this paper. In this approach, models are sampled randomly from the set of all possible models, according to the probability of the models given the training data. Our work applies the random sampling approach to probabilistic classifiers by computing an approximation to the posterior model distribution given the training data, and generating committee members from that distribution. McCallum and Nigam (1998) use a similar approach for sample selection on text categorization using a naive Bayes classifier. The primary difference is that they skew example selection using *density-weighted sampling*, such that documents that are similar to many other documents in the training set will be selected for labeling with a higher probability.

Matan (1995) presents two other methods for random sampling. In the first method, he trains committee members on different subsets of the training data. In his second method, for neural network models, Matan generates committee members by backpropagation training using different initial weights in the networks so that they reach different local minima. A similar approach is taken by Liere and Tadepalli (1997), who applied a committee-based selection approach to text categorization using the Winnow learning algorithm (Littlestone, 1988) which learns linear classifiers. They represented the model space by a set of classifiers (the *model set*). Each classifier in the model set learns independently from labeled examples, having been initialized with a different initial hypothesis (thus at any point the set gives a selection of the possible hypotheses given the training data). Labeling decisions are performed based on two models chosen at random from the model set. If the models disagree on a document's class, the document's label is requested, and all models in the space are updated.

### 2.1 Query By Committee

As mentioned above, this paper follows theoretical work on sample selection in the Query By Committee (QBC) paradigm (Seung et al., 1992; Freund et al., 1997). This method was proposed for learning binary (non-probabilistic) concepts in cases where there exists a prior probability distribution measure over the concept class. QBC selects 'informative' training examples out of a stream of unlabeled examples. When an example is selected the learner queries the teacher for its correct label and adds it to the training set. As examples are selected for training, they restrict the set of *consistent concepts*, i.e, the set of concepts that label all the training examples correctly (the version space).

A simple version of QBC, which was analyzed by Freund et al. (1997) (see also the summary in Freund, 1994), uses the following selection algorithm:

1. Draw an unlabeled input example at random from the probability distribution of the example space.
2. Select at random two hypotheses according to the prior probability distribution of the concept class, restricted to the set of consistent concepts.
3. Select the example for training if the two hypotheses disagree on its classification.

Freund et al. prove that, under some assumptions, this algorithm achieves an exponential reduction in the number of labeled examples required to achieve a desired classification accuracy, compared with random selection of training examples. This speedup is achieved because the algorithm tends to select examples that split the version space into two parts of similar size. One of these parts is eliminated from the version space after the example and its correct label are added to the training set.

### 2.2 Selection For Probabilistic Classifiers

We address here the problem of sample selection for training a probabilistic classifier. Classification in this framework is performed by a probabilistic model which, given an input example, assigns a probability (or a probability-based score) to each possible classification and selects the best classification. Probabilistic classifiers do not fall within the framework addressed in the theoretical QBC work. Training a probabilistic classifier involves estimating the values of model parameters which determine a probability estimate for each possible classification of an example. While we expect that in most cases the optimal classifier will assign the highest probability to the correct class, this is not guaranteed to always occur. Accordingly, the notion of a consistent hypothesis is generally not applicable to probabilistic classifiers. Thus, the posterior distribution over classifiers given the training data cannot be defined as the restriction of the prior to the set of consistent hypotheses. Rather, within a Bayesian framework, the posterior distribution is defined by the statistics of the training set, assigning higher probability to those classifiers which are more likely given the statistics.

We now discuss some desired properties of examples that are selected for training. Generally speaking, a training example contributes data to several statistics, which in turn determine the estimates of several parameter values. An informative example is therefore one whose contribution to the statistics leads to a useful improvement of parameter estimates. Assuming the existence of an optimal classification model for the given concept (such as a maximum likelihood model), we identify three properties of parameters for which acquiring additional statistics is most beneficial:

1. The current estimate of the parameter is uncertain due to insufficient statistics in the training set. An uncertain estimate is likely to be far from the true value of the parameter and can cause incorrect classification. Additional statistics would bring the estimate closer to the true value.

2. Classification is sensitive to changes in the current estimate of the parameter. Otherwise, acquiring additional statistics is unlikely to affect classification and is therefore not beneficial.

3. The parameter takes part in calculating class probabilities for a large proportion of examples. Parameters that are only relevant for classifying few examples, as determined by the probability distribution of the input examples, have low utility for future estimation.

The committee-based selection scheme, as we describe further below, tends to select examples that affect parameters with the above three properties. Property 1 is addressed by randomly picking parameter values for committee members from the posterior distribution of parameter estimates (given the current statistics). When the statistics for a parameter are insufficient the variance of the posterior distribution of the estimates is large, and hence there will be large differences in the values of the parameter picked for different committee members. Note that property 1 is not addressed when uncertainty in classification is only judged relative to a *single* model (as in, e.g., Lewis & Gale, 1994). Such an approach captures uncertainty with respect to given parameter values, in the sense of property 2, but it does not model uncertainty about the choice of these values in the first place (the use of a single model is criticized by Cohn et al., 1994).

Property 2 is addressed by selecting examples for which committee members highly disagree in classification. Thus, the algorithm tends to acquire statistics where uncertainty in parameter estimates entails uncertainty in actual classification (this is analogous to splitting the version space in QBC). Finally, property 3 is addressed by independently examining input examples which are drawn from the input distribution. In this way, we implicitly model the expected utility of the statistics in classifying future examples.

### 2.3 Paper Outline

The following section defines the basic concepts and notation that we will use in the rest of the paper. Section 4 presents a general selection scheme along with variant selection algorithms. The next two sections demonstrate the effectiveness of the sample selection scheme. Section 5 presents results on an artificial "colorful coin flipper" problem, providing a simple illustration of the operation of the proposed system. Section 6 presents results for the task of stochastic part-of-speech tagging, demonstrating the usefulness of committee-based sample selection in the real world.

## 3. Definitions

The concern of this paper is how to minimize the number of labeled examples needed to learn a classifier which accurately classifies input examples $e$ by classes $c \in C$, where $C$ is a known set of possible classes. During learning, a stream of unlabeled examples is supplied for free, with examples drawn from an unknown probability distribution. There is a cost, however, for the learning algorithm to obtain the true label of any given example. Our objective is to reduce this cost as much as possible, while still learning an accurate classifier.

We address the specific case of *probabilistic classifiers*, where classification is done on the basis of a score function, $F_M(c, e)$, which assigns a score to each possible class of an input example. The classifier assigns the input example to the class with the highest score. $F_M$ is determined by a probabilistic model $M$. In many applications, $F_M$ is the conditional probability function, $P_M(c|e)$, specifying the probability of each class given the example. Alternatively, other score functions that denote the likelihood of the class may be used (such as an odds ratio). The particular type of model used for classification determines the specific form of the score, as a function of features of the example.

A probabilistic model $M$, and thus the score function $F_M$, is defined by a set of parameters, $\{\alpha_i\}$, giving the probabilities of various possible events. For example, a model for part-of-speech tagging contains parameters such as the probability of a particular word being a verb or a noun. During training, the values of the parameters are estimated from a set of statistics, $S$, extracted from a training set of labeled examples. A particular model is denoted by $M = \{a_i\}$, where each $a_i$ is a specific value for the corresponding $\alpha_i$.

## 4. Committee-Based Sample Selection

This section describes the algorithms which apply the *committee-based* approach for evaluating classification uncertainty of each input example. The learning algorithm evaluates an example by giving it to a *committee* containing several versions, or copies, of the classifier, all 'consistent' with the training data seen so far. The greater the agreement of the committee members on the classification of the example, the greater our certainty in its classification. This is because if the training data entails a specific classification with high certainty, then most (in a probabilistic sense) versions of the classifier consistent with the data will produce that classification. An example is selected for labeling, therefore, when the committee members disagree on its appropriate classification.

### 4.1 Generating A Committee

To generate a committee with $k$ members, we randomly choose $k$ models according to the posterior distribution $P(M|S)$ of possible models given the current training statistics. How this sampling is performed depends on the form of this distribution, which in turn depends on the form of the model. Thus when implementing committee-based selection for a particular problem, an appropriate sampling procedure must be devised. As an illustration of committee generation, the rest of this section describes the sampling process for models consisting of independent binomial parameters or multinomial parameter groups.

Consider first a model containing a single binomial parameter $\alpha$ (the probability of a success), with estimated value $a$. The statistics $S$ for such a model are given by $N$, the number of trials, and $x$, the number of successes in those trials.

Given $N$ and $x$, the 'best' model parameter value can be estimated by any of several estimation methods. For example, the maximum likelihood estimate (MLE) for $\alpha$ is $a = \frac{x}{N}$, giving the model $M = \{\alpha = \frac{x}{N}\}$. When generating a committee of models, however, we are not interested in the 'best' model, but rather in sampling the distribution of models given the statistics. For our example, we need to sample the posterior density of estimates for $\alpha$, namely $p(\alpha = a|S)$. In the binomial case, this density is the *beta distribution* (Johnson, 1970). Sampling this distribution yields a set of estimates scattered around $\frac{x}{N}$ (assuming a uniform prior), where the variance of these estimates gets smaller as $N$ gets larger. Each such estimate participates in a different member of the committee. Thus, the more statistics there are for estimating the parameter, the closer are the estimates used by different models in the committee.

Now consider a model consisting of a single group of interdependent parameters defining a multinomial. In this case, the posterior is a Dirichlet distribution (Johnson, 1972). Committee members are generated by sampling from this joint distribution, giving values for all the model parameters.

For models consisting of a set of independent binomials or multinomials, sampling $P(M|S)$ amounts to sampling each of the parameters independently. For models with more complex dependencies among parameters sampling may be more difficult. In practice, though, it may be possible to make enough independence assumptions to make sampling feasible.

Sampling the posterior generates committee members whose parameter estimates differ most when they are based on low training counts and tend to agree when based on high counts. If the classification of an example relies on parameters whose estimates by com-

For each unlabeled input example $e$:

1. Draw 2 models randomly from $P(M|S)$, where $S$ are statistics acquired from previously labeled examples;
2. Classify $e$ by each model, giving classifications $c_1$ and $c_2$;
3. If $c_1 \neq c_2$, select $e$ for annotation;
4. If $e$ is selected, get its correct label and update $S$ accordingly.

Figure 1: The two member sequential selection algorithm.

mittee members differ, *and* these differences affect classification, then the example would be selected for learning. This leads to selecting examples which contribute statistics to currently unreliable estimates that also have an effect on classification. Thus we address Properties 1 and 2 discussed in Section 2.2.

## 4.2 Selection Algorithms

Within the committee-based paradigm there exist different methods for selecting informative examples. Previous research in sample selection has used either *sequential* selection (Seung et al., 1992; Freund et al., 1997; Dagan & Engelson, 1995), or *batch* selection (Lewis & Catlett, 1994; Lewis & Gale, 1994). We present here general algorithms for both sequential and batch committee-based selection. In all cases, we assume that before any selection algorithm is applied a small amount of labeled *initial training* is supplied, in order to initialize the training statistics.

### 4.2.1 Two Member Sequential Selection

*Sequential* selection examines unlabeled examples as they are supplied, one by one, and estimates their expected information gain. Those examples determined to be sufficiently informative are selected for training. Most simply, we can choose a committee of size two from the posterior distribution of the models, and select an example when the two models disagree on its classification. This gives the parameter-free, *two member sequential selection algorithm*, shown in Figure 1. This basic algorithm has no parameters.

### 4.2.2 General Sequential Selection

A more general selection algorithm results from:

- Using a larger number $k$ of committee members, in order to evaluate example informativeness more precisely,
- More refined example selection criteria, and

For each unlabeled input example $e$:

1. Draw $k$ models $\{M_i\}$ randomly from $P(M|S)$ (possibly using a temperature $t$);
2. Classify $e$ by each model $M_i$ giving classifications $\{c_i\}$;
3. Measure the disagreement $D(e)$ based on $\{c_i\}$;
4. Decide whether or not to select $e$ for annotation, based on the value of $D(e)$;
5. If $e$ is selected, get its correct label and update $S$ accordingly.

Figure 2: The general sequential selection algorithm.

- Tuning the frequency of selection by replacing $P(M|S)$ with a distribution with a different variance. This has the effect of adjusting the variability among the committee members chosen. In many cases (eg., HMMs, as described in Section 6 below) this can be implemented by a parameter $t$ (called the *temperature*), used as a multiplier of the variance of the posterior parameter distribution.

This gives the *general sequential selection algorithm*, shown in Figure 2.

It is easy to see that two member sequential selection is a special case of general sequential selection. In order to instantiate the general algorithm for larger committees, we need to fix a general measure $D(e)$ for disagreement (step 3), and a decision method for selecting examples according to this disagreement (step 4).

We measure disagreement by the entropy of the distribution of classifications 'voted for' by the committee members. This *vote entropy* is a natural measure for quantifying the uniformity of classes assigned to an example by the different committee members[1]. We further normalize this entropy by a bound on its maximum possible value ($\log \min(k, |c|)$), giving a value between 0 and 1. Denoting the number of committee members assigning a class $c$ for input example $e$ by $V(c, e)$, the *normalized vote entropy* is:

$$D(e) = -\frac{1}{\log \min(k, |C|)} \sum_c \frac{V(c,e)}{k} \log \frac{V(c,e)}{k}$$

Normalized vote entropy has the value one when all committee members disagree, and the value zero when they all agree, taking on intermediate values in cases with partial agreement.

We consider here two alternatives for the selection criterion (step 4). The simplest is *thresholded selection*, in which an example is selected for annotation if its normalized vote entropy exceeds some threshold $\theta$. Another alternative is *randomized selection*, in which an example is selected for annotation based on the flip of a coin biased according to the vote entropy—a higher vote entropy corresponding to a higher probability of selection. We

1. McCallum and Nigam (1998) have suggested an alternative measure, the *KL-divergence to the mean* (Pereira, Tishby, & Lee, 1993). It is not clear whether that measure has an advantage over the simpler entropy function.

For a batch $B$ of $N$ examples:

1. For each example $e$ in $B$:
   (a) Draw $k$ models randomly from $P(M|S)$;
   (b) Classify $e$ by each model, giving classifications $\{c_i\}$;
   (c) Measure the disagreement $D(e)$ for $e$ based on $\{c_i\}$;
2. Select for annotation the $m$ examples with the highest $D(e)$;
3. Update $S$ by the statistics of the selected examples.

Figure 3: The batch selection algorithm.

use a simple model where the selection probability is a linear function of normalized vote entropy: $P(e) = gD(e)$, calling $g$ the *entropy gain*[2].

#### 4.2.3 Batch Selection

An alternative to sequential selection is *batch selection.* Rather than evaluating examples individually for their informativeness a large batch of $N$ examples is examined, and the $m$ best are selected for annotation. The *batch selection algorithm* is given in Figure 3.

This procedure is repeated sequentially for successive batches of $N$ examples, returning to the start of the corpus at the end. If $N$ is equal to the size of the corpus, batch selection selects the $m$ globally best examples in the corpus at each stage (as in Lewis & Catlett, 1994). Batch selection has certain theoretical drawbacks (Freund et al., 1997), particularly that it does not consider the distribution of input examples. However, as shown by McCallum and Nigam (1998), the distribution of the input examples can be modeled and taken into account during selection. They do this by combining their disagreement measure with a measure of example density, which produces good results with batch selection (this work is discussed in more detail below in Section 7.2). A separate difficulty with batch selection is that it has the computational disadvantage that it must look at a large number of examples before selecting any. As the batch size is decreased, batch selection behaves similarly to sequential selection.

## 5. Example: Colorful Coin Flipper

As an illustrative example of a learning task, we define a *colorful coin-flipper* (CCF) as a machine which contains an infinite number of coins of various colors. The machine chooses coins to flip, one by one, where each color of coin has a fixed (unknown) probability of being chosen. When a coin is flipped, it comes up heads with probability determined solely by its color. Before it flips a coin, the machine tells the learner which color of coin it has chosen to

2. The selection method used in (Dagan & Engelson, 1995) is randomized sequential selection using this linear selection probability model, with parameters $k$, $t$ and $g$.

flip. In order to know the outcome of the flip, however, the learner must pay the machine. In training, the learner may choose the colors of coins whose outcomes it will examine. The objective of selective sampling is to choose so as to minimize the training cost (number of flips examined) required to attain a given prediction accuracy for flip outcomes.

For the case of the CCF, an example $e$ is a coin flip, characterized by its color, and its class $c$ is either heads or tails. Note that we do not require that flips of a given color always have the same class. Therefore the best that we can hope to do is classify according to the most likely class for each color.

For a CCF, we can define a model whose parameters are the heads probabilities for the coins of each particular color. So, for a CCF with three colors, one possible model would be $m = \{r = 0.8, g = 0.66, b = 0.2\}$, giving the probabilities of heads for red, green, and blue coins, respectively. A coin of a given color will then be classified 'heads' if its score (given directly by the appropriate model parameter) is $> \frac{1}{2}$, and 'tails' otherwise.

### 5.1 Implementation Of Sample Selection

Training a model for a CCF amounts to counting the proportion of heads for each color, providing estimates of heads probabilities. In *complete training* every coin flip in the training sequence is examined and added to the counts. In sample selection we seek to label and count only training flips of those colors for which additional counts are likely to improve the model's accuracy. Useful colors to train on are either those for which few training examples have so far been seen, or those whose current probability estimates are near 0.5 (cf. Section 2.2).

Recall that for sample selection we build a committee by sampling models from $P(M|S)$. In the case of a CCF, all of the model parameters $\alpha_i$ (the heads probabilities for different colors) are independent, and so sampling from $P(M|S)$ amounts to sampling independently for each of the parameters.

While the form of the posterior distribution $P(\alpha_i = a_i|S)$ is given by the beta distribution, we found it technically easier to use a normal approximation, which was found satisfactory in practice. Let $N_i$ be the number of coin flips of color $i$ seen so far, and $n_i$ be the number of those flips which came up heads. We approximate $P(\alpha_i = a_i|S)$ as a truncated normal distribution (restricted to [0,1]), with estimated mean $\mu_i = \frac{n_i}{N_i}$ and variance $\sigma_i^2 = \frac{\mu_i(1-\mu_i)}{N_i}$. This approximation made it easy to also incorporate a 'temperature' parameter $t$ (as in Section 4.2.2), which is used as a multiplier for the variance estimate $\sigma_i^2$. Thus, we actually approximate $P(\alpha_i = a_i|S)$ as a truncated normal distribution with mean $\mu_i$ and variance $\sigma_i^2 t$. Sampling from this distribution was done using the algorithm given by Press, Flannery, Teukolsky, and Vetterling (1988) for sampling from a normal distribution.

### 5.2 Vote Entropy

The CCF is useful to illustrate the importance of determining classification uncertainty using the vote entropy over a committee of models rather than using the entropy of the class distribution given by a single model (as discussed in Section 2). Consider a CCF with

| Model | Red | Blue | Green |
|---|---|---|---|
| 0 | 0.55 (heads) | 0.45 (tail) | 0.48 (heads) |
| 1 | 0.55 (heads) | 0.45 (tail) | 0.75 (tail) |
| 2 | 0.60 (heads) | 0.55 (heads) | 0.85 (tail) |
| 3 | 0.60 (heads) | 0.55 (heads) | 0.95 (tail) |

(a)

| Color | $D(e)$ | ACDE |
|---|---|---|
| Red | 0.0 | 0.98 |
| Blue | 1.0 | 0.99 |
| Green | 0.81 | 0.68 |

(b)

Figure 4: (a) A committee of CCF models. (b) The resultant vote entropy for each color.

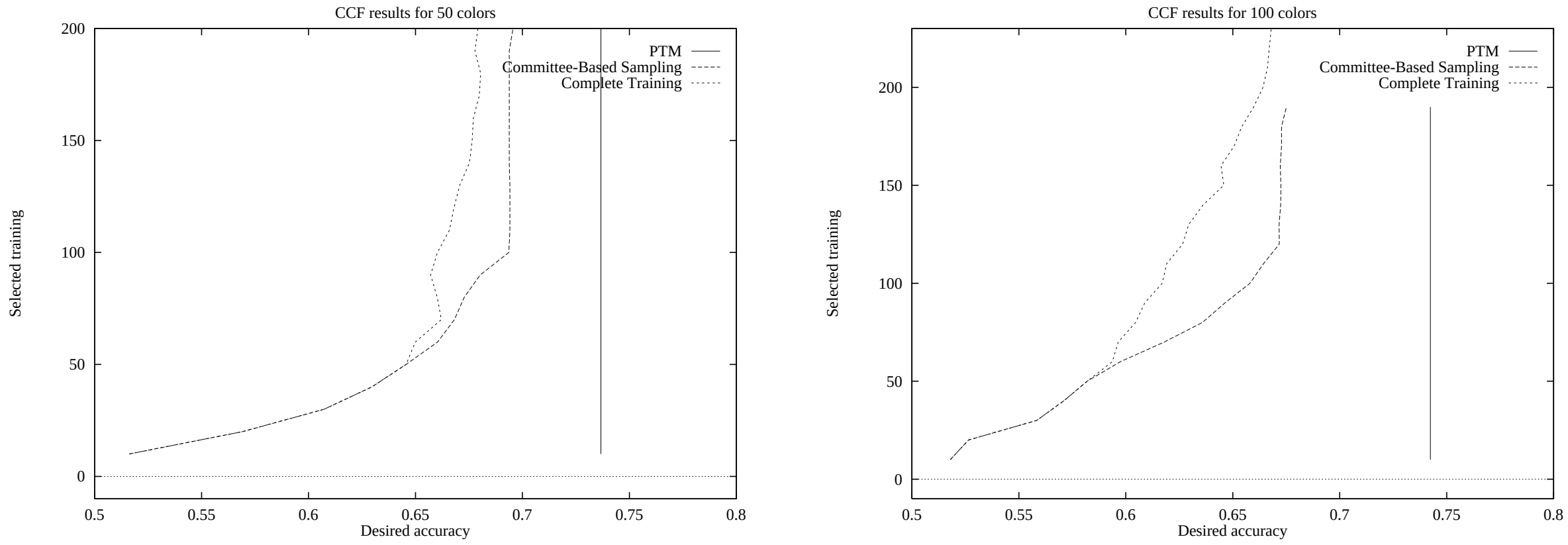

(a) (b)

Figure 5: CCF results for random CCFs with 50 and 100 different coin colors. Results are averaged over 4 different such CCFs, comparing complete training with two member sample selection. The figures show the amount of training required for a desired classification accuracy: (a) for 50 colors, (b) for 100 colors.

three coin colors, red, blue, and green. Suppose the 4-member committee in Figure 4(a) is generated. From this committee, we estimate for each color its vote entropy $D(e)$, as well as the average of the class distribution entropies given by each of the individual models (ACDE), given in Figure 4(b).

If we compare the entropies of red and blue, for example, we see that their entropies over the expected *class probability* distribution are both quite high (since both estimated class probabilities are near 0.5). However, when we consider their vote entropies (over the *assigned classes*), blue has maximal entropy, since the range of possible models straddles a class boundary (0.5), while red has minimal entropy, since the range of possible models does not straddle a class boundary. That is, it is quite certain that the optimal classification for red is "heads". We also see how green has a higher vote entropy than red, although its average class distribution entropy is lower. This shows the importance of using vote entropy for selection.

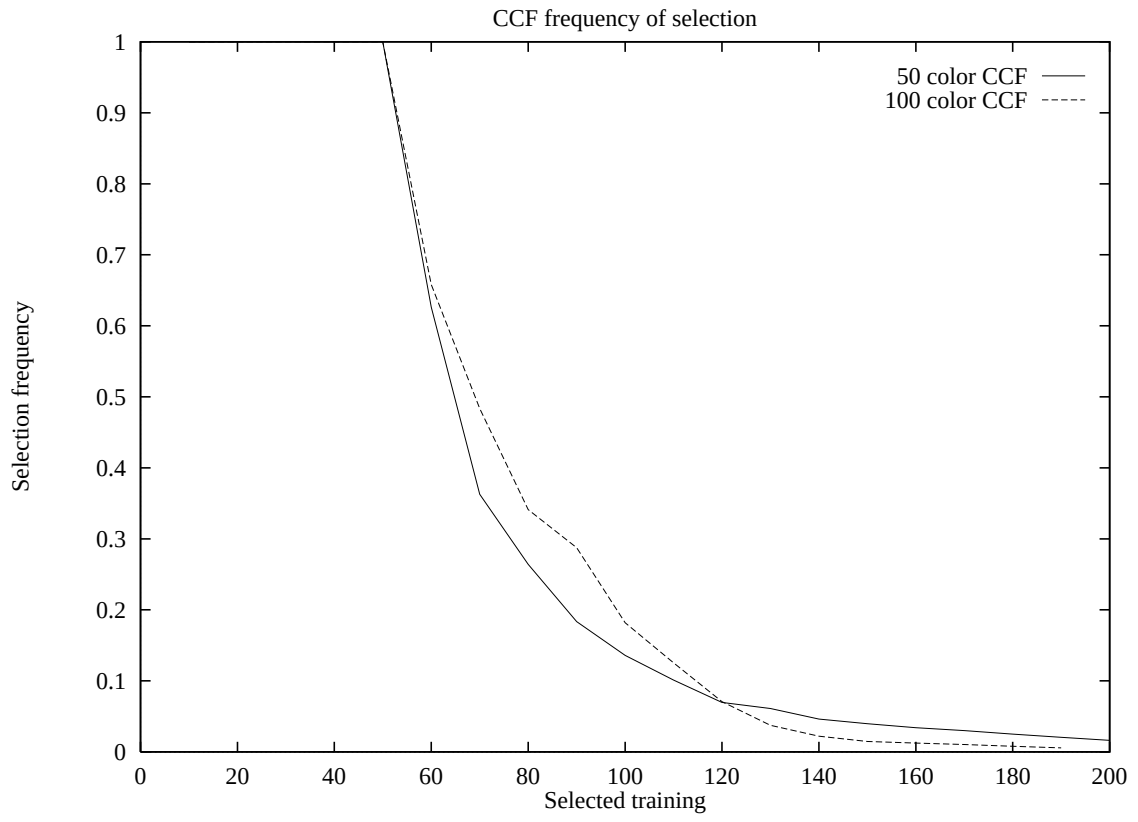

Figure 6: Frequency of selection vs. amount of selected training for CCFs with 50 and 100 colors, averaged for 4 different CCFs.

### 5.3 Results

We simulated sample selection for the simple CCF model in order to illustrate some of its properties. In the following, we generated random CCFs with a fixed number of coins by randomly generating occurrence probabilities and heads probabilities for each coin color. We then generated learning curves for complete training, on all input examples, and for two member sample selection, using 50 coin-flips for initial training. Both complete training and sample selection were run on the same coin-flip sequences. Accuracy was measured by computing the expected accuracy (assuming an infinite test set) of the MLE model generated by the selected training. The figures also show the accuracy for the theoretical *perfectly trained model* (PTM) which knows all of the parameters perfectly.

Figure 5 summarizes the average results for 4 comparison runs of complete vs. sample selection for CCFs of 50 and 100 coins. In Figures 5(a) and (b), we compare the amount of selected training required to reach a given desired accuracy. We see in both cases that as soon as sample selection starts operating, its efficiency is higher than complete training, and the gap increases in size as greater accuracy is desired. In Figure 6, we examine the cumulative frequency of selection (ratio between the number of selected examples and the total number of examples seen) as learning progresses. We see here an exponential decrease in the frequency of selection, as expected in the case of QBC for non-probabilistic models (analyzed in Seung et al., 1992; Freund et al., 1997).

## 6. Application: Stochastic Part-Of-Speech Tagging

We have applied committee-based selection to the real-world task of learning Hidden Markov Models (HMMs) for part-of-speech tagging of English sentences. Part-of-speech tagging is the task of labeling each word in the sentence with its appropriate part of speech (for example, labeling an occurrence of the word 'hand' as a noun or a verb). This task is non-trivial since determining a word's part of speech depends on its linguistic context. HMMs have been used extensively for this task (e.g., Church, 1988; Merialdo, 1991), in most cases trained from corpora which have been manually annotated with the correct part of speech for each word.

### 6.1 HMMs And Part-Of-Speech Tagging

A first-order Hidden Markov Model (HMM) is a probabilistic finite-state string generator (Rabiner, 1989), defined as a set of states $Q = \{q_i\}$, a set of output symbols $\Sigma$, a set of *transition probabilities* $P(q_i \rightarrow q_j)$ of each possible transition between states $q_i$ and $q_j$, a set of *output probabilities* $P(a|q)$ for each state $q$ to output each symbol $a \in \Sigma$, and a distinguished *start state* $q_0$. The probability of a string $s = a_1 a_2 \cdots a_n$ being generated by an HMM is given by

$$\sum_{q_1 \cdots q_n \in Q^n} \left( \prod_{i=1}^{n} P(q_{i-1} \rightarrow q_i) P(a_i | q_i) \right) \quad ,$$

the sum, for all paths through the HMM, of the joint probability that the path was traversed and that it output the given string. In contrast with ordinary Markov Models, in an HMM it is not known which sequence of states generated a given string (hence the term 'hidden').

HMMs have been used widely in speech and language processing. In particular, an HMM can be used to provide a classification model for sequence elements: If we need to classify each element in a sequence, we encode each possible class by a state in an HMM. Training the HMM amounts to estimating the values of the transition and output probabilities. Then, given a sequence for classification, we assume that it was generated by the HMM and compute the most likely state sequence for the string, using the Viterbi algorithm[3] (Viterbi, 1967).

An HMM can be used for part-of-speech tagging of words by encoding each possible part-of-speech tag, $t$ (noun, verb, adjective, etc.), as an HMM state. The output probabilities, $P(w|t)$, give the probability of producing each word $w$ in the language conditioned on the current tag $t$. The transition probabilities, $P(t_1 \rightarrow t_2)$, give the probability of generating a word with the tag $t_2$ given that the previous word's tag is $t_1$. This constitutes a weak syntactic model of the language. This model is often termed the *tag-bigram model*[4].

Given an input word sequence $W = w_1 \cdots w_n$, we seek the most likely tag sequence $T = t_1 \cdots t_n$:

$$\begin{aligned} \arg\max_T P(T|W) &= \arg\max_T \frac{P(T,W)}{P(W)} \\ &= \arg\max_T P(T,W) \end{aligned}$$

3. An alternative classification scheme is to compute the most likely state for each individual element (instead of the most likely state sequence) by the Forward-Backward algorithm (Rabiner, 1989) (also called the Baum-Welch algorithm Baum, 1972). We do not address here this alternative, which is computationally more expensive and is typically not used for part-of-speech tagging. It is possible, however, to apply the committee-based selection method also for this type of classification.
4. It should be noted that practical implementations of part-of-speech tagging often employ a tag-trigram model, in which the probability of a tag depends on the last two tags rather than just the last one. The committee-based selection method which we apply here to the bigram model can easily be applied also to the trigram case.

since $P(W)$ is a constant. Thus we seek the $T$ which maximizes

$$P(T,W) = \prod_{i=1}^{n} P(t_{i-1}{\rightarrow}t_i)P(w_i|t_i)$$

For technical convenience, we use Bayes' theorem to replace each $P(w_i|t_i)$ term by the term $\frac{P(t_i|w_i)P(w_i)}{P(t_i)}$, noting that $P(w_i)$ does not effect the maximization over tag sequences and can therefore be omitted (following Church, 1988). The parameters of a part-of-speech model, then, are: *tag probabilities* $P(t_i)$, *transition probabilities* $P(t_{i-1}{\rightarrow}t_i)$, and *lexical probabilities* $P(t|w)$.

Supervised training of the tagger is performed using a tagged corpus (text collection), which was manually labeled with the correct part-of-speech for each word. Maximum likelihood estimates (MLEs) for the parameters are easily computed from word and tag counts from the corpus. For example, the MLE of $P(t)$ is the fraction of tag occurrences in the corpus that were the tag $t$, whereas $P(t|w)$ is the ratio between the count for the word $w$ being labeled with the tag $t$ and the total count for $w$. In our committee-based selection scheme, the counts are used also to compute the posterior distributions for parameter estimates, as discussed below in Section 6.2.

We next describe the application of our committee-based selection scheme to the HMM classification framework. First we will discuss how to sample from the posterior distributions over the HMM parameters $P(t_i{\rightarrow}t_j)$ and $P(t|w)$, given training statistics.[5] We then discuss the question of how to define an example for training—an HMM deals with (in principle) infinite strings; on what substrings do we make decisions about labeling? Finally, we describe how to measure the amount of disagreement between committee members.

### 6.2 Posterior Distributions For Multinomial Parameters

In this section, we consider how to select committee members based on the posterior parameter distributions $P(\alpha_i = a_i|S)$ for an HMM, assuming a uniform prior. First note that the parameters of an HMM define a set of multinomial probability distributions. Each multinomial corresponds to a conditioning event and its values are given by the corresponding set of conditioned events. For example, a transition probability parameter $P(t_i{\rightarrow}t_j)$ has conditioning event $t_i$ and conditioned event $t_j$.

Let $\{u_i\}$ denote the set of possible values of a given multinomial variable (e.g., the possible tags for a given word), and let $S = \{n_i\}$ denote a set of statistics extracted from the training set, where $n_i$ is the number of times that the value $u_i$ appears in the training set. We denote the total number of appearances of the multinomial variable as $N = \sum_i n_i$. The parameters whose distributions we wish to estimate are $\alpha_i = P(u_i)$.

The maximum likelihood estimate for each of the multinomial's distribution parameters, $\alpha_i$, is $\hat{\alpha}_i = \frac{n_i}{N}$. In practice, this estimator is usually smoothed in some way to compensate for data sparseness. Such smoothing typically reduces the estimates for values with positive

5. We do not sample the model space over the tag probability parameters, since the amount of data for tag frequencies is large enough to make their MLEs quite definite.

counts and gives small positive estimates for values with a zero count. For simplicity, we first describe here the approximation of $P(\alpha_i = a_i|S)$ for the unsmoothed estimator[6].

The posterior $P(\alpha_i = a_i|S)$ is a Dirichlet distribution (Johnson, 1972); for ease of implementation, we used a generalization of the normal approximation described above (Section 5.1) for binomial parameters. We assume first that a multinomial is a collection of independent binomials, each of which corresponds to a single value $u_i$ of the multinomial; we then separately apply the constraint that the parameters of all these binomials should sum to 1. For each such binomial, we sample from the approximate distribution (possibly with a temperature $t$). Then, to generate a particular multinomial distribution, we renormalize the sampled parameters so they sum to 1.

To sample for the smoothed estimator, we first note that the estimator for the smoothed model (interpolating with the uniform) is

$$\hat{\alpha}_i^S = \frac{(1-\lambda)n_i + \lambda}{(1-\lambda)N + \lambda\nu} \quad ,$$

where $\lambda \ll 1$ is a smoothing parameter controlling the amount of smoothing (in our experiments $\lambda = 0.05$), and $\nu$ is the number of possible values for the given multinomial. We then sample for each $i$ from the truncated normal approximation (as in Section 5) for the *smoothed* estimate, i.e, with mean $\mu = \hat{\alpha}_i^S$ and variance $\sigma^2 = \frac{\mu(1-\mu)}{N}$. Normalization for the multinomial is then applied as above.

Finally, to generate a random HMM given statistics $S$, we note that all of its parameters $P(t_i \rightarrow t_j)$ and $P(t|w)$ are independent of each other. We thus independently choose values for the HMM's parameters from each multinomial distribution.

### 6.3 Examples For HMM Training

Typically, concept learning problems are formulated such that there is a set of training examples that are independent of each other. When training an HMM, however, each state/output pair is dependent on the previous state, so we are presented (in principle) with a single infinite input string for training. In order to perform sample selection, we must divide this infinite string into (short) finite strings.

For part-of-speech tagging, this problem may be solved by considering each sentence as an individual example. More generally, we can break the text at any point where tagging is unambiguous. In particular, it is common to have a *lexicon* which specifies which parts-of-speech are possible for each word (i.e, which of the parameters $P(t|w)$ are positive). In bigram tagging, we can use unambiguous words (those with only one possible part of speech) as example boundaries. Similar natural breakpoints occur in other HMM applications; for example, in speech recognition we can consider different utterances separately. In other cases of HMM learning, where such natural breakpoints do not occur, some heuristic will have to be applied, preferring to break at 'almost unambiguous' points in the input.

6. In the implementation we smooth the MLE by interpolation with a uniform probability distribution, following Merialdo (1991). Adaptation of $P(\alpha_i = a_i|S)$ to the smoothed version of the estimator is given below.

### 6.4 Quantifying Disagreement

Recall that our selection algorithms decide whether or not to select an example based on how much the committee members disagree on its labeling. As discussed in Section 4.2.2, we suggest the use of vote entropy for measuring classification disagreement between committee members. This idea is supported by the fact that we found empirically that the average normalized vote entropy for words which the tagger (after some training) classified correctly was 0.25, whereas the average entropy for incorrectly classified words was 0.66. This demonstrates that vote entropy is a useful measure of classification uncertainty (likelihood of error) based on the training data.

In bigram tagging, each example consists of a sequence of several words. In our implementation, we measured vote entropy separately for each word in the sequence, and use the average vote entropy over the sequence as our measurement of disagreement for the example. We use the average entropy rather than the entropy over the entire sequence, because the number of committee members is small with respect to the total number of possible tag sequences.

### 6.5 Results

We now present our results on applying committee-based sample selection to bigram part-of-speech tagging, comparing it with complete training on all examples in the corpus. Evaluation was performed using the University of Pennsylvania tagged corpus from the ACL/DCI CD-ROM I. For ease of implementation, we used a complete (closed) lexicon which contains all the words in the corpus.[7] Approximately 63% of the word occurrences in the corpus are ambiguous in the lexicon (have more than one possible part-of-speech).

Each committee-based selection algorithm was initialized using the first 1,000 words from the corpus, and then examined the following examples in the corpus for possible labeling. The training set consisted of the first million words in the corpus, with sentence ordering randomized to compensate for inhomogeneity in corpus composition. The test set was a separate portion of the corpus consisting of 20,000 words, starting just after the first 1,000,000.

We compared the amount of training required by different selection methods to achieve a given tagging accuracy on the test set, where both the amount of training and tagging accuracy are measured over ambiguous words.[8]

#### 6.5.1 Labeling Efficiency

7. We use the lexicon provided with Brill's part-of-speech tagger (Brill, 1992). While in an actual application a complete lexicon would not be available, our results using a complete lexicon are valid, as the evaluation of complete training and committee-based selection is comparative.
8. Most other work on tagging has measured accuracy over all words, not just ambiguous ones. Complete training of our system on 1,000,000 words gave us an accuracy of 93.5% over ambiguous words, which corresponds to an accuracy of 95.9% over all words in the test set, comparable to other published results on bigram tagging.

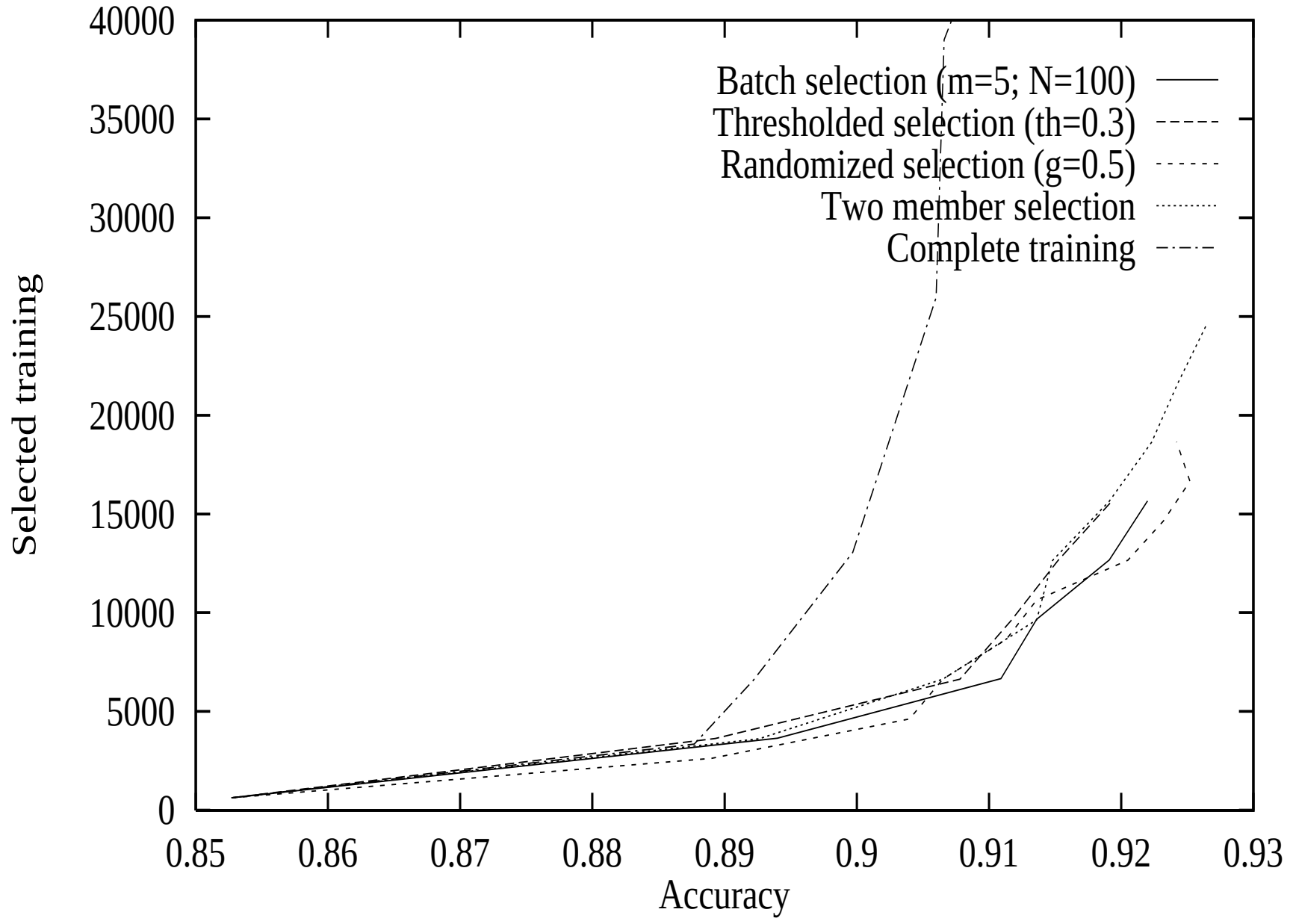

Figure 7: Labeled training versus classification accuracy. In batch, random, and thresholded runs, $k = 5$ and $t = 50$.

Figure 7 presents a comparison of the results of several selection methods. The reported parameter settings are the best found for each selection method by manual tuning. Figure 7 shows the advantage that sample selection gives with regard to annotation cost. For example, complete training requires annotated examples containing 98,000 ambiguous words to achieve a 92.6% accuracy, while the selection methods require only 18,000–25,000 ambiguous words to achieve this accuracy. We also find that, to a first approximation, all of the methods considered give similar results. Thus, it seems that a refined choice of the selection method is not crucial for achieving large reductions in annotation cost.

### 6.5.2 Computational Efficiency

Figure 8 plots classification accuracy versus number of words *examined*, instead of those *selected*. Complete training is clearly the most efficient in these terms, as it learns from all examples examined. The selective methods are similar, though two member selection seems to require somewhat fewer examples for examination than the other methods. Furthermore, since only two committee members are used this method is computationally more efficient in evaluating each examined example.

### 6.5.3 Model Size

The ability of committee-based selection to focus on the more informative parts of the training corpus is analyzed in Figure 9. Here we examined the number of lexical and bigram counts that were stored (i.e, were non-zero) during training, using the two member selection algorithm and complete training. As the graphs show, committee-based selection achieves the same accuracy as complete training with fewer lexical and bigram counts. To achieve 92% accuracy, two member selection requires just 6200 lexical counts and 750 bigram counts, as compared with 15,800 lexical counts and 1100 bigram counts for complete training. This implies that many counts in the data are not needed for correct tagging, since smoothing estimates the probabilities equally well.[9] Committee-based selection ignores these counts, focusing its efforts on parameters which improve the model's performance. This behavior has an additional practical advantage of reducing the size of the model significantly. Also, the average count is lower in a model constructed by selective training than in a fully trained model, suggesting that the selection method tends to avoid using examples which increase the counts for already known parameters.

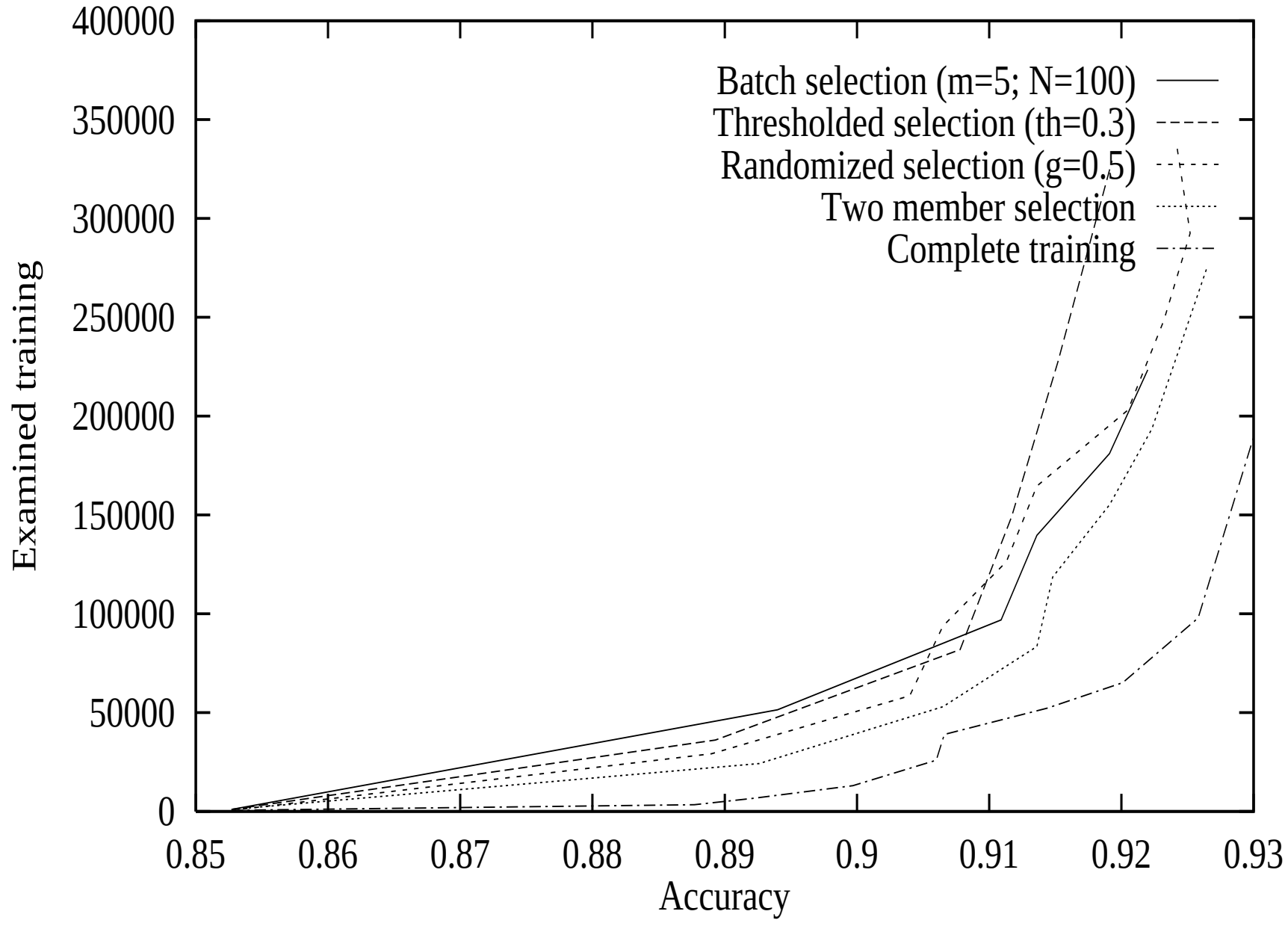

Figure 8: Examined training (both labeled and unlabeled) versus classification accuracy. In batch, random, and thresholded runs, $k = 5$ and $t = 50$.

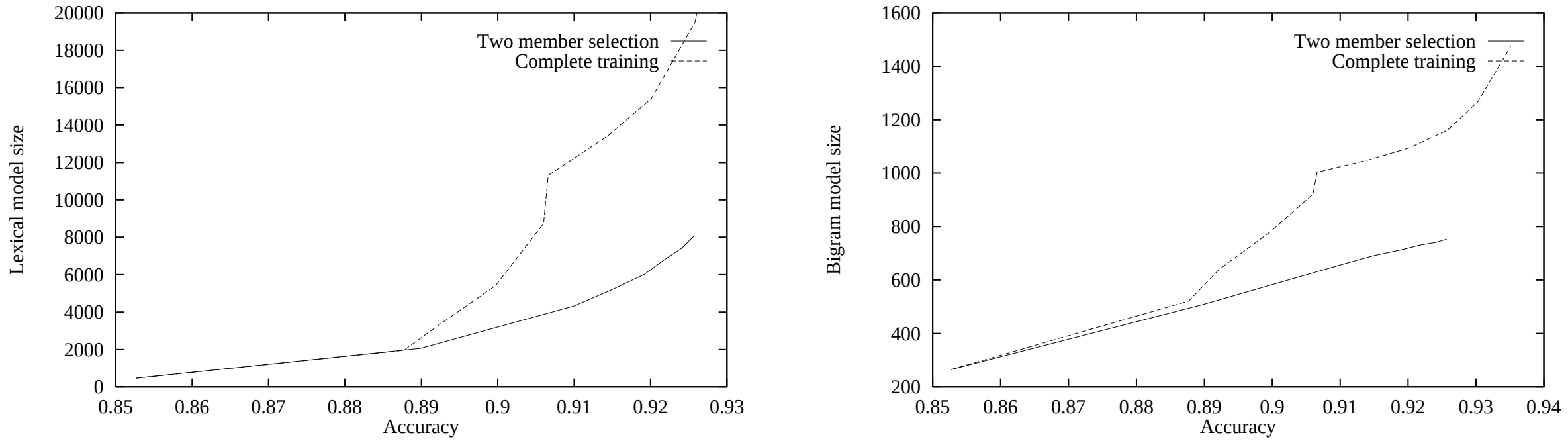

(a) (b)

Figure 9: Numbers of frequency counts $> 0$, plotted ($y$-axis) versus classification accuracy ($x$-axis). (a) Lexical counts ($\text{freq}(t, w)$) (b) Bigram counts ($\text{freq}(t_1 \rightarrow t_2)$).

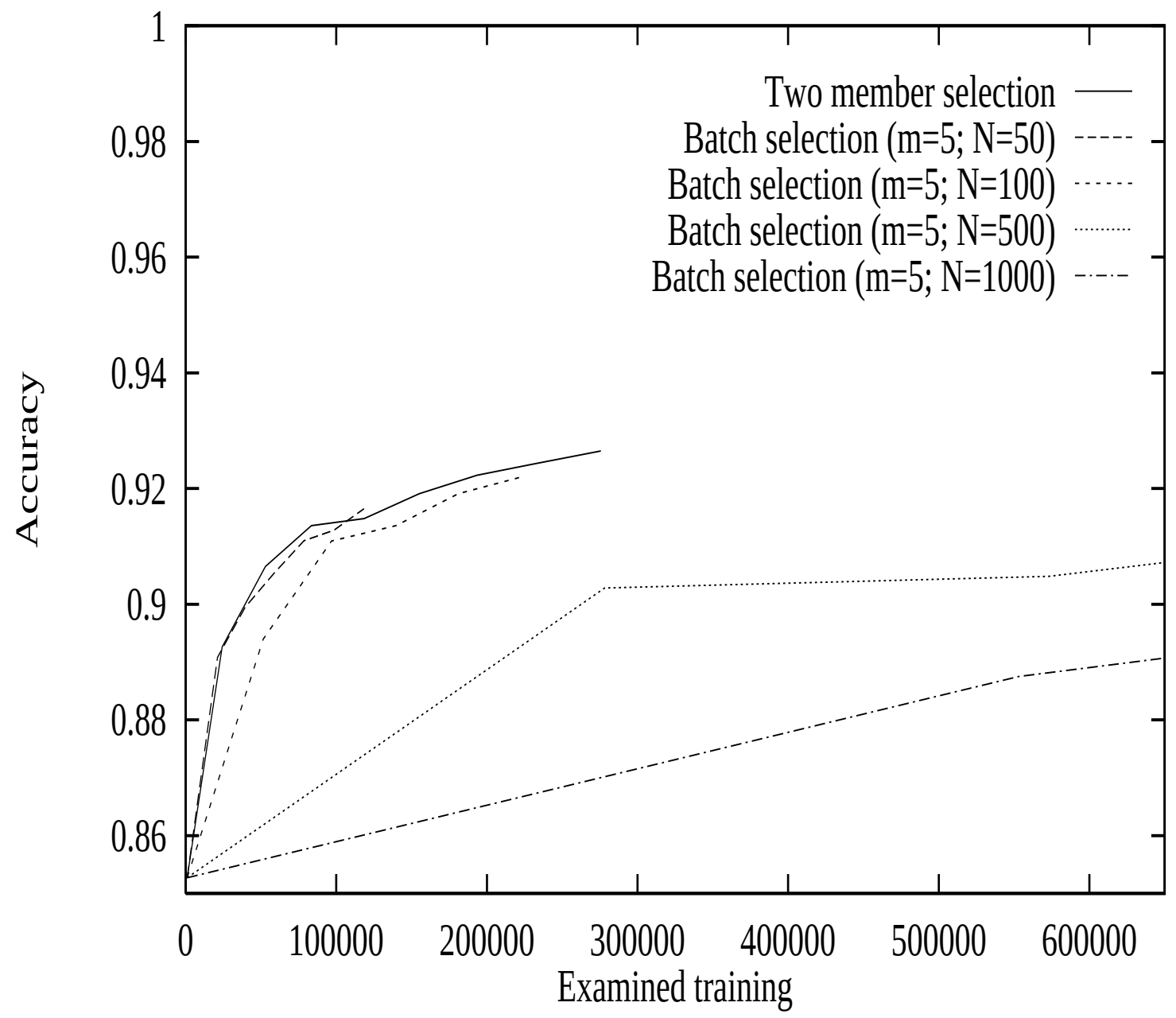

Figure 10: Evaluating batch selection, for $m = 5$. Classification accuracy versus number of words examined from the corpus for different batch sizes.

#### 6.5.4 Batch Selection

We investigated the properties of batch selection, varying batch size from 50 to 1000 examples, fixing the number of examples selected from each batch at 5. We found that in terms of the number of labeled examples required to attain a given accuracy, selection for these different batch sizes performed similarly. This means that increased batch size

9. As mentioned above, in the tagging phase we smooth the MLE estimates by interpolation with a uniform probability distribution, following Merialdo (1994).

does not seem to improve the effectiveness of selection. On the other hand, we did not see a decrease in performance with increased batch size, which we might have expected due to poorer modeling of the input distribution (as noted in Section 2.2). This may indicate that even a batch size of 1000 (selecting just 1/200 of the examples seen) is small enough to let us model the input distribution with reasonable accuracy. However, the similarity in performance of the different batch sizes to each other and to sequential selection does not hold with respect to the amount of *unlabeled* training used. Figure 10 shows accuracy attained as a function of the amount of unlabeled training used. We see quite clearly that, as expected, using larger batch sizes required examining a far larger number of unlabeled training examples in order to obtain the same accuracy.

## 7. Discussion

### 7.1 Committee-Based Selection As A Monte-Carlo Technique

We can view committee-based selection as a Monte-Carlo method for estimating the probability distribution of classes assigned to an example over all possible models, given the training data. The proportion of votes among committee members for a class $c$ on an example $e$ is a sample-based estimate of the probability, for a model chosen randomly from the posterior model distribution, of assigning $c$ to $e$. That is, the the proportion of votes for $c$ given $e$, $\frac{V(c,e)}{k}$, is a Monte-Carlo estimate of

$$P^*(c|e,S) = \int_{\mathcal{M}} T_M(c|e)P(M|S)dM$$

where $M$ ranges over possible models (vectors of parameter values) in the model space $\mathcal{M}$, $P(M|S)$ is the posterior probability density of model $M$ given statistics $S$, and $T_M(c|e) = 1$ if $c$ is the highest probability class for $e$ based on $M$ (i.e, if $c = \arg\max_{c_i} P_M(c_i|e)$, where $P_M(c|e)$ is the class probability distribution for $e$ given by model $M$), and 0 otherwise. Vote entropy, as defined in Section 4.2.2, is thus an approximation of the entropy of $P^*$. This entropy is a direct measure of uncertainty in example classification over the possible models.

Note that we measure entropy over the *final classes* assigned to an example by possible models (i.e, $T_M$), not over the *class probabilities* given by a single model (i.e, $P_M$), as illustrated by the CCF example of Section 5.2. Measuring entropy over $P_M$ (say, by looking at the expected probability over all models) would not properly address properties 1 and 2 discussed in Section 2.2.

### 7.2 Batch Selection

Property 3 discussed in Section 2.2 states that parameters that affect only few examples have low overall utility, and so atypical examples are not very useful for learning. In sequential selection, this property is addressed by independently examining input examples which are drawn from the input distribution. In this way, we implicitly model the distribution of model parameters used for classifying input examples. Such modeling, however, is not inherent in the basic form of batch selection, which can lead to it being less effective (Freund et al., 1997).

This difficulty of batch selection is addressed directly by McCallum and Nigam (1998), who describe a version of batch selection (called *pool-based sampling*), which differs from the basic batch selection scheme presented in Section 4.2.3 in two ways. First, they quantify disagreement between committee members by the *KL-divergence to the mean* (Pereira et al., 1993), rather than vote entropy. More significantly, their disagreement measure is combined with an explicit density measure in *density-weighted sampling*, such that documents that are similar to many other documents in the training set will be more probably selected for labeling. This is intended to address property 3 in Section 2.2. The authors found empirically that for text classification using naive Bayes, their density-weighted pool-based selection method using KL-divergence to the mean improved learning efficiency over complete training. They also found that sequential selection using vote entropy was worse than complete training for their problem.

We hypothesize that this is due to the high degree of sparseness of the example space (text documents), which leads to a large proportion of the examples being atypical (even though documents similar to a given atypical document are rare, many different atypical documents occur.) Since this is the case, the sequential variant may tend to select many atypical documents for labeling, which would degrade learner performance by skewing the statistics. This problem can be remedied by adding density-weighting to sequential selection in future research. This may yield an efficient sequential selection algorithm that also works well in highly sparse domains.

## 8. Conclusions

Labeling large training sets for supervised classification is often a costly process, especially for complicated domain areas such as natural language processing. We have presented an approach for reducing this cost significantly using committee-based sample selection, which reduces redundant annotation of examples that contribute little new information. The method is applicable whenever it is possible to estimate a posterior distribution over the model space given the training data. We have shown how to apply it to training Hidden Markov Models, and demonstrated its effectiveness for the complex task of part of speech tagging. Implicit modeling of uncertainty makes the selection system generally applicable and relatively simple to implement. In practical settings, the method may be applied in a semi-interactive process, in which the system selects several new examples for annotation at a time and updates its statistics after receiving their labels from the user.

The committee-based sampling method addresses the three factors which relate the informativeness of a training example to the model parameters that it affects. These factors are: (1) the statistical significance of the parameter's estimate, (2) the parameter's effect on classification, and (3) the probability that the parameter will be used for classification in the future. The use of a committee models the uncertainty in classification relative to the entire model space, while sequential selection implicitly models the distribution of the examples.

Our experimental study of variants of the selection method suggests several practical conclusions. First, it was found that the simplest version of the committee-based method, using a two-member committee, yields reduction in annotation cost comparable to that of the multi-member committee. The two-member version is simpler to implement, has no parameters to tune and is computationally more efficient. Second, we generalized the selection scheme giving several alternatives for optimizing the method for a specific task. For bigram tagging, comparative evaluation of the different variants of the method showed similar large reductions in annotation cost, suggesting the robustness of the committee-based approach. Third, sequential selection, which implicitly models the expected utility of an example relative to the example distribution, worked in general better than batch selection. Recent results on improving batch selection by modeling explicitly the 'typicality' of examples suggest further comparison of the two approaches (as discussed in the previous section). Finally, we studied the effect of sample selection on the size of the trained model, showing a significant reduction in model size for selectively trained models.

In future research we propose to investigate the applicability and effectiveness of committee-based sample selection for additional probabilistic classification tasks. Furthermore, the generality obtained by implicitly modeling information gain suggests using variants of committee-based sampling also in non-probabilistic contexts, where explicit modeling of information gain may be impossible. In such contexts, committee members might be generated by randomly varying some of the decisions made in the learning algorithm.

### Acknowledgments

Discussions with Yoav Freund, Yishai Mansour, and Wray Buntine greatly enhanced this work. The first author was at Bar-Ilan University while this work was performed, and was supported by the Fulbright Foundation during part of the work.